%% file: root.tex
\documentclass[letterpaper, 10 pt, conference]{ieeeconf}
\IEEEoverridecommandlockouts
\usepackage{cite}
\usepackage{amsmath,amssymb,amsfonts}
\usepackage{algorithmic}
\usepackage{graphicx,adjustbox}
\usepackage{textcomp}
\usepackage{xcolor}
\usepackage{color}
\usepackage{afterpage}
\usepackage{makecell} 

\usepackage{colortbl}
\usepackage{float}
\usepackage[font=small,labelfont=bf]{caption}
\usepackage{wrapfig,lipsum,booktabs}
\usepackage{subfig}
\usepackage{units}
\def\BibTeX{{\rm B\kern-.05em{\sc i\kern-.025em b}\kern-.08em
    T\kern-.1667em\lower.7ex\hbox{E}\kern-.125emX}}
\usepackage{multirow}
\newcolumntype{P}[1]{>{\centering\arraybackslash}p{#1}}
\newcolumntype{M}[1]{>{\centering\arraybackslash}m{#1}}

\begin{document}

\definecolor{antiquefuchsia}{rgb}{0.57, 0.36, 0.51}
\definecolor{teal}{rgb}{0.196, 0.659, 0.620}
\definecolor{TEE_blue}{RGB}{0,128,255}
\definecolor{nicepurple}{RGB}{150,50,200}

\newcommand{\MFR}[1]{\textcolor{nicepurple}{[MFR: #1]}}
\newcommand{\TEE}[1]{\textcolor{TEE_blue}{[TEE: #1]}}
\newcommand{\RJW}[1]{\textcolor{red}{[RJW: #1]}}
\newcommand{\AK}[1]{\textcolor{cyan}{[AK: #1]}}
\newcommand{\sas}[1]{\textcolor{orange}{[SS: #1]}}

\newcommand{\MES}[1]{\textcolor{teal}{[MES: #1]}}
\newcommand{\DSE}[1]{\textcolor{olive}{[DSE: #1]}}

\title{A Supervised Autonomous Resection and Retraction Framework for Transurethral Enucleation of the Prostatic Median Lobe}

\author{Mariana Smith$^{*1}$, Tanner Watts$^{*2}$, Susheela Sharma Stern$^{3}$, Brendan Burkhart$^{1}$, Hao Li$^{4}$, \\ Alejandro O. Chara$^{5}$, Nithesh Kumar$^{3}$, James Ferguson$^{6}$, Ayberk Acar$^{4}$, Jesse F. d'Almeida$^{3}$, \\ Lauren Branscombe$^{7}$, Lauren Shepard$^{8}$, Ahmed Ghazi$^{8}$, Ipek Oguz$^{4}$, Jie Ying Wu$^{4}$, \\ Robert J. Webster III$^{3}$, Axel Krieger$^{1}$, and Alan Kuntz$^{6}$%
\thanks{$^{*}$ Both authors contributed equally to this work.}
\thanks{$^{1}$Mariana Smith, Brendan Burkhart, and Axel Krieger are with the Laboratory for Computational Sensing and Robotics, Johns Hopkins University, Baltimore, MD, 21211, USA. (email: { msmit458}@jh.edu)}%
\thanks{$^{2}$Tanner Watts is with the Department of Electrical and Computer Engineering at Vanderbilt University, Nashville, TN, 37203, USA. }
\thanks{$^{3}$Susheela Sharma Stern, Nithesh Kumar, Jesse F. d'Almeida, and Robert J. Webster III are with the Department of Mechanical Engineering, Vanderbilt University, Nashville, TN, 37203, USA. }
\thanks{$^{4}$Hao Li, Ayberk Acar, Ipek Oguz, and Jie Ying Wu are with the Department of Computer Science, Vanderbilt University, Nashville, TN, 37203, USA. }
\thanks{$^{5}$Alejandro Chara is with the Department of General Surgery, Vanderbilt University Medical Center, Nashville, TN, 37203, USA.}
\thanks{$^{6}$Alan Kuntz is with the Robotics Center and the Kahlert School of Computing at the University of Utah, Salt Lake City, UT 84112, USA.}
\thanks{$^{7}$Lauren Branscombe is with Virtuoso Surgical, Nashville, TN, USA}
\thanks{$^{8}$Lauren Shepard and Ahmed Ghazi are with the Department of Urology, Johns Hopkins University James Buchanan Brady Urological Institute, Baltimore, Maryland, USA.}
\thanks{Research reported in this publication was supported by the Advanced Research Projects Agency for Health (ARPA-H) under Award Number D24AC00415 for ALISS. The ARPA-H award provided 95\% of total costs and total up to \$11,338,286.10 and nongovernmental funds 5\% of total cost and \$596,751.90 nongovernmental funds. The contents are those of the authors. They may not reflect the policies of the Department of Health and Human Services or the U.S. government. The content is solely the responsibility of the authors and does not necessarily represent the official views of the Advanced Research Projects Agency for Health. This material is also supported in part by the NSF Foundational Research in Robotics (FRR) Faculty Early Career Development Program (CAREER) under grant number 2144348. Any opinions, findings, and conclusions or recommendations expressed in this material are those of the authors and do not necessarily reflect the views of NSF.
}%
}


\maketitle

\begin{abstract}

Concentric tube robots (CTRs) offer dexterous motion at millimeter scales, enabling minimally invasive procedures through natural orifices. 
This work presents a coordinated model-based resection planner and learning-based retraction network that work together to enable semi-autonomous tissue resection using a dual-arm transurethral concentric tube robot (the Virtuoso).
The resection planner operates directly on segmented CT volumes of prostate phantoms, automatically generating tool trajectories for a three-phase median lobe resection workflow: left/median trough resection, right/median trough resection, and median blunt dissection. 
The retraction network, PushCVAE, trained on surgeon demonstrations, generates retractions according to the procedural phase. 
The procedure is executed under Level-3 (supervised) autonomy on a prostate phantom composed of hydrogel materials that replicate the mechanical and cutting properties of tissue. 
As a feasibility study, we demonstrate that our combined autonomous system achieves a 97.1\% resection of the targeted volume of the median lobe. 
Our study establishes a foundation for image-guided autonomy in transurethral robotic surgery and represents a first step toward fully automated minimally-invasive prostate enucleation.

\end{abstract}

\input{01-Introduction}
\input{02-Methods}

\input{03-Experiments}
\input{04-Conclusion}

\bibliographystyle{IEEEtran}
\bibliography{ProstateRefs}

\end{document}

%% file: 01-Introduction.tex
\section{Introduction}

Benign Prostatic Hyperplasia (BPH) affects roughly 210 million men worldwide~\cite{Vos2012GBD} and is the most common symptomatic condition in aging males~\cite{Issa2006Prevalence}. About one in three men will require surgery for BPH during their lifetime~\cite{Glynn1985BPH}, with over 300,000 procedures performed annually in the United States alone~\cite{McConnell1994BPH}. These numbers are projected to nearly double over the next decade as the population ages~\cite{Collins2014TURP}.

Prostate enucleation, which removes the internal adenoma while leaving the outer capsule intact, has been shown in numerous clinical trials to outperform conventional Transurethral Resection of the Prostate (TURP) with lower complication rates, lower re-operation rates, and even lower costs~\cite{Giannantonio2011ARFI, Gupta2006BPH, Kuntz2004HoLEP, Montorsi2004HoLEP}. However, it is rarely performed~\cite{Bhojani2014BPH} due to the higher technical difficulty associated with the procedure. Surgeons must manipulate a straight tool through an endoscope to precisely remove prostatic tissue to the level of the underlying capsule. Intra-operative visualization and navigation are extremely challenging in such a confined workspace~\cite{ElHakim2002HoLEP, Kim2013HoLEP, Placer2009HoLEP, Seki2003HoLEP, Shah2007HoLEP}.

Transurethral robotic systems such as the Virtuoso Endoscopic System (VES) (Virtuoso Surgical, Nashville, TN, USA) can overcome these challenges to make enucleation more accessible. The Virtuoso system integrates two concentric tube robot (CTR) manipulators and monocular visualization into an endoscopic form factor of only 12 mm diameter, enabling highly-dexterous minimally-invasive transurethral access. CTRs are made from telescoping, pre-curved nitinol tubes which curve and extend when rotated and translated, producing tentacle-like motions~\cite{Mahoney2018CTRReview, Gilbert2016CTR, BurgnerKahrs2015Continuum, Mitros2022CTR, Nwafor2023DesignCTR, Dupont2022Continuum, Wang2022Eccentric}. Even when manufactured at submillimeter diameters to fit through standard endoscopes, CTRs maintain high dexterity for surgical tasks in organs such as the prostate~\cite{Hendrick2015Handheld, Amanov2020Transurethral, dAlmeida2025LowField}, trachea~\cite{Gafford2020Bronchoscopy}, and uterus\cite{Harvey2020Hysteroscopy}. The VES's compact design also enables exceptional relative positioning accuracy between its camera and instruments (located only millimeters away from one another), in contrast to competitors like the da Vinci system (Intuitive, Sunnyvale, CA, USA), which have much longer kinematic chains and lower accuracy. This precision paves the way for robotic automation of surgical subtasks in minimally-invasive transurethral procedures, such as resection and retraction for prostate enucleation.

In this paper, we present a first-of-its-kind system for supervised autonomous transurethral resection of the median prostate lobe using the VES. Our approach integrates CT-guided preoperative resection planning with a learning-based retraction network, applied to novel hydrogel prostate phantoms which look, feel, and cut like real tissues ~\cite{Li2015SiliconePVA, Melnyk2020NephrectomySim, Witthaus2020CRPMS}. The workflow mirrors surgeon demonstrations of enucleation using the VES system, with a three-phase approach: left/median trough resection, right/median trough resection, and blunt median lobe dissection. We demonstrate that our integrated pipeline can perform a complete median lobe resection under Level 3 (supervised) autonomy~\cite{Yang2017}, representing the first integration of model-based planning and learning-based manipulation in a minimally invasive robotic prostate surgery platform.

%% file: 02-Methods.tex
\section{Methods}

\subsection{System Hardware}
\begin{figure}
    \centering
    \includegraphics[width = \linewidth]{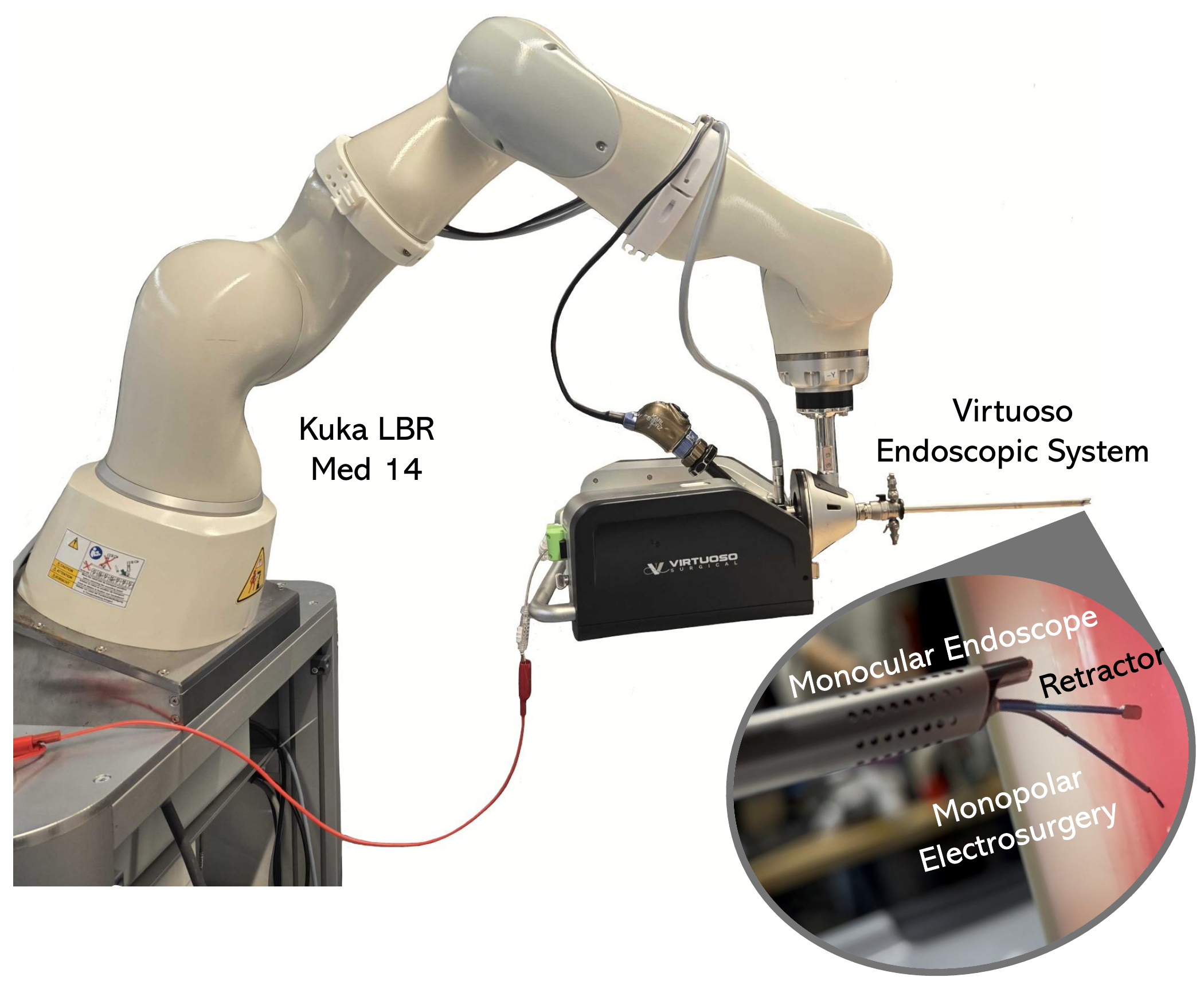}
    \caption{Robotic system for experimentation, comprised of a Kuka LBR Med 14 for gross positioning and a VES mounted to its end-effector. The VES consists of a monocular endoscope with two CTR tools: one spatula retractor and one monopolar electrosurgery probe.}
    \label{fig:robot}
\end{figure}

The hardware configuration for this work (Fig. \ref{fig:robot}) consists of a Kuka LBR Med 14 (Kuka, Augsburg, Germany) with a VES (Virtuoso Surgical, Nashville, TN, USA) mounted to its end-effector. The VES has a monocular endoscope (Karl Storz, Tuttlingen, Germany) and two CTR manipulators which exit through discrete channels under the endoscope. The left CTR has a spatula-shaped retractor tool at its tip, while the right CTR has a monopolar electrosurgery probe inserted through its lumen. The back end of the electrosurgery probe is wired to an electrosurgical generator nearby, which is set to a power level of 80W and activated on/off over ROS2. Due to the workspace constraints of the CTRs (each tool having a workspace of diameter 40mm), the Kuka must be used for gross positioning of the endoscope and tools within the anatomy. Both the Kuka and VES are equipped with ROS2 interfacing to enable automated control. Each of the two CTR arms is comprised of two tubes, a curved outer tube and a straight inner tube. Each arm has 4 degrees of freedom (DOF) corresponding to translation and rotation of each tube, enabling complex motion in 3D. The Kuka can be jogged in Cartesian position or velocity control, with a Remote Center of Motion (RCM) enforced at the phantom's transurethral entry port. We used known CAD dimensions and endoscope specifications to derive the transformation from the Kuka's tool frame to the VES's endoscope. A hand-eye calibration obtained the transformation from one CTR base frame to the endoscope. Known CAD dimensions derived the transformation between the two CTRs. Thus, the full transformation tree was complete, enabling execution of globally planned trajectories with intra-operative gross positioning.

\subsection{Phantom Setup}
\begin{figure}[h!]
    \centering
    \includegraphics[width = \linewidth]{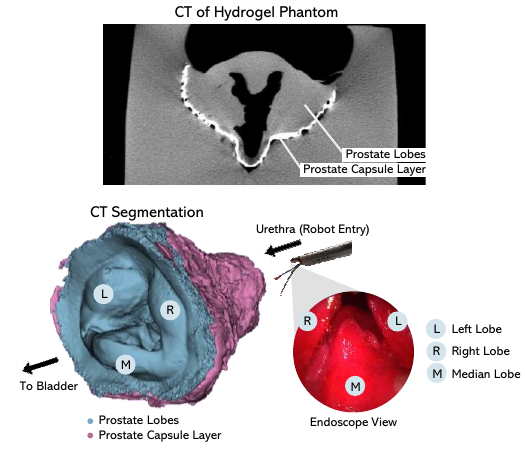}
    \caption{A CT scan is taken of the hydrogel prostate phantom, which allows segmentation of the capsule and lobes using embedded contrast agent. The robot enters the model through the urethra, where it can visualize the left, right, and median lobes of the prostate in its monocular endoscope.}
    \label{fig:phantom}
\end{figure}

The goal of a prostate enucleation is to resect the lobes of the prostate down to, but not past, the surface of the prostatic capsule. Thus, our model-based resection planner requires a 3D volume of the prostate (the inner lobe surface and outer capsule surface) as its input. Currently, patients undergoing prostate enucleation procedures commonly receive an MRI before operation, in which the boundary of the capsule layer is clearly visible to the surgeon. Due to equipment limitations, we were unable to collect MRIs of our phantoms; therefore, we used CT to obtain the same 3D information. The capsule layer is not visible in CT, like it is in MRI, so we embedded a Barium Sulfate contrast agent in the capsule layer of the phantom for visibility. The resulting CT (Fig. \ref{fig:phantom}) was segmented into a 3D volume of the lobe tissue within the capsule boundary, ready to be passed to our automated resection planning module.

\subsection{Model-Based Resection Planner}
\begin{figure}
    \centering
    \includegraphics[width = \linewidth]{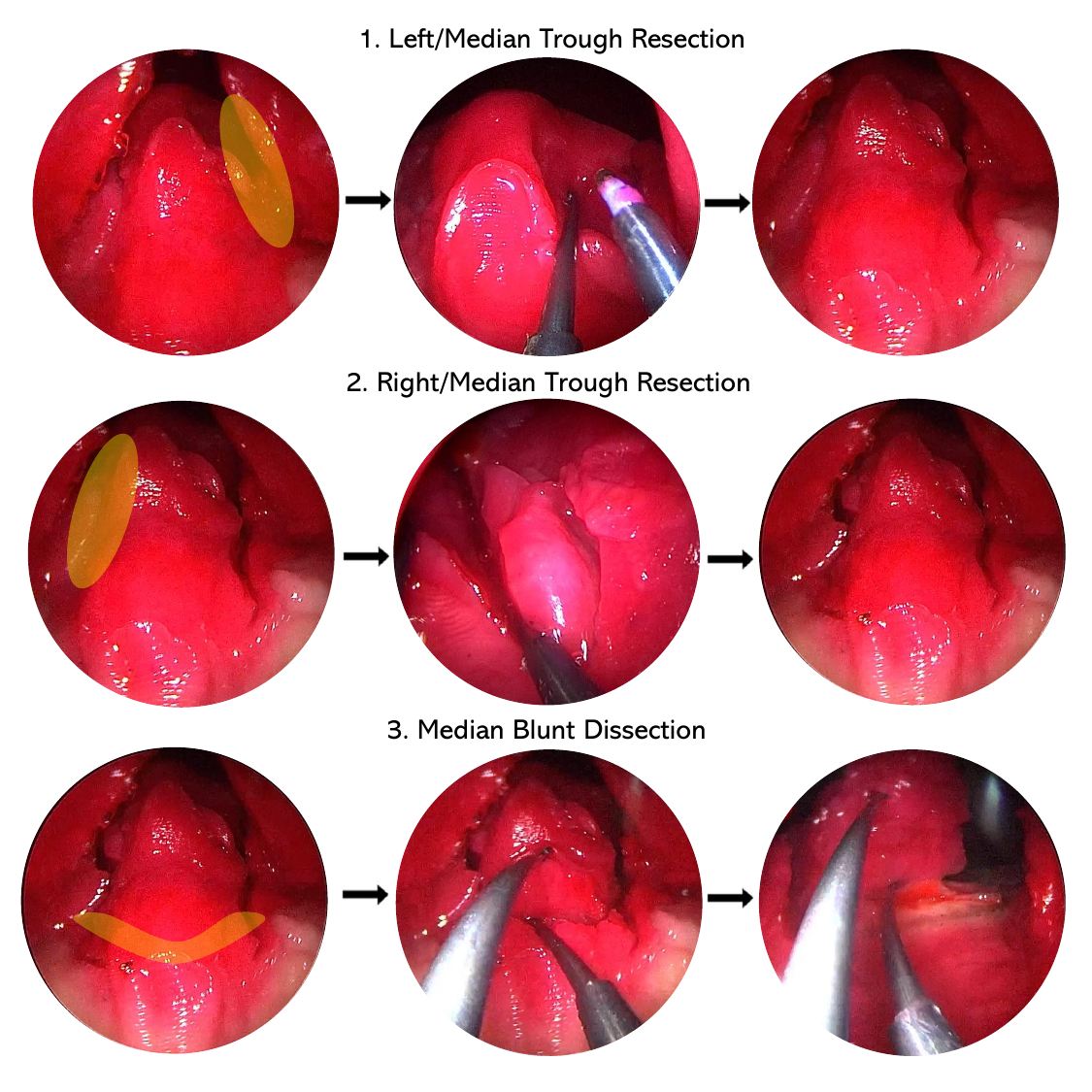}
    \caption{Minimally-invasive robotic procedural workflow for resection of the median prostate lobe using the VES, demonstrated here in teleoperation by a human operator.}
    \label{fig:teleopsteps}
\end{figure}

\begin{figure*}[h!]
    \centering
    \includegraphics[width = \textwidth]{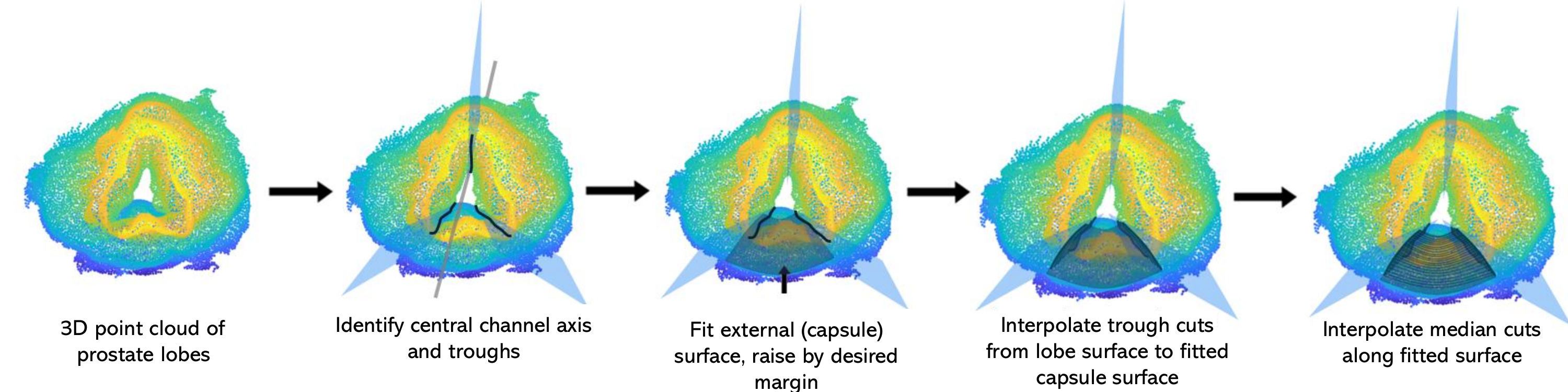}
    \caption{Process diagram of the model-based resection planner.}
    \label{fig:planner}
\end{figure*}

The full prostate enucleation procedure consists of the complete removal of the left, right, and median lobes, removing all the internal prostate tissue and leaving the outer capsule. In this work, we automated median lobe removal as an initial step toward full enucleation automation. We break the median lobe resection into three phases, based on surgeon demonstrations of this procedure using the VES (Fig. \ref{fig:teleopsteps}). Phase 1 is the left/median trough resection, in which the surgeon detaches the left and median lobes, yielding a trough between them. Phase 2 is the right/median trough resection, in which the surgeon cuts the right and median lobes apart, yielding a trough between them. Phase 3 is the median blunt dissection, in which the surgeon gradually resects the median lobe along its border with the capsule surface, essentially connecting the two troughs that were just created in Phases 1 and 2. Phase 3 ends when the median lobe is fully disconnected from the bulk tissue. We set out to automate this procedure using the same three-phase workflow as the human surgeon. The high-level overview of our automatic model-based planner is illustrated in Fig. \ref{fig:planner}.

\begin{figure}[h!]
    \centering
    \includegraphics[width = 0.5\linewidth]{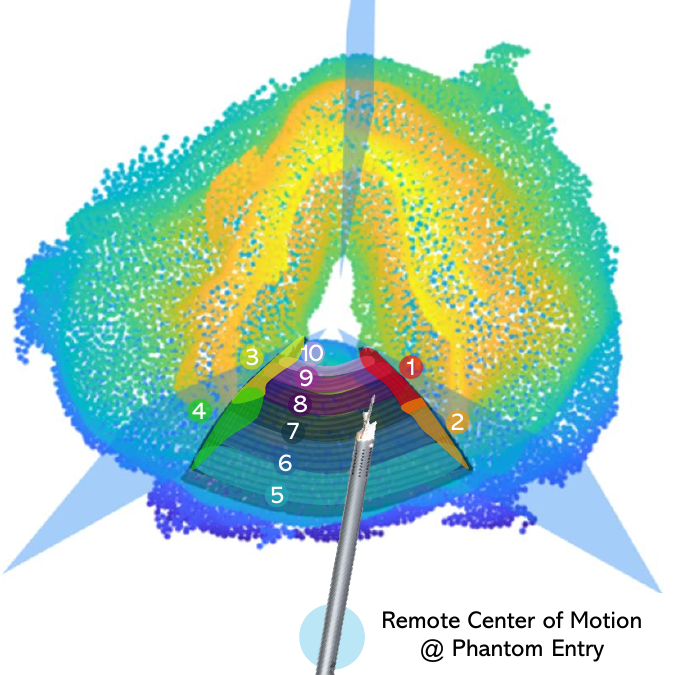}
    \caption{Groupings of planned cuts to execute at discrete global positions of the endoscope, necessary to ensure reachability. Global movements are executed manually with the Kuka robot, with an RCM constraint at the entry.}
    \label{fig:global}
\end{figure}

\textit{Identification of Central Channel:} The input point cloud is downsampled to 10,000 vertices. We estimate the central channel axis in three stages. First, rays are uniformly sampled from the capsule centroid, keeping only those avoiding collisions with capsule or lobe surfaces. Surviving directions form two clusters corresponding to opposite channel orientations; PCA identifies the principal direction for each, and the orientation maximizing capsule cross-section is selected. Second, the axis position is optimized by sampling radial translations orthogonal to the axis and selecting the candidate with maximal clearance from lobe points.

\textit{Identification of Troughs:} Troughs are detected in two stages. Rays are cast radially along the channel axis to identify lobe-surface points, forming local “triangles.” Aggregated points are clustered into three troughs, and PCA fits a line to each. Troughs are labeled top/center, left, and right using the instrument Y-axis and the channel axis to define a reference plane consistent with the endoscopic view.

\textit{Capsule Surface Fit:} Within the left–right plane corridor, capsule points are selected and transformed into a local UVW frame. A 5th-order polynomial $W=f(U,V)$ is fit via linear regression. The surface is evaluated on a UV grid, transformed back to world coordinates, raised by the resection margin along $W$, and cropped between the left and right planes to generate the 3D capsule surface for trajectory planning.

\textit{Interpolated Cuts:} For each trough, lobe points are snapped to the trough plane, sorted along the channel axis, and smoothed. Matching capsule tracks are extracted and clipped at their first intersection. Intermediate layers are linearly interpolated at 2mm spacing, regularized for spacing, downsampled, and augmented with reach-in waypoints. Median cuts are generated similarly using the UVW frame.

\textit{Grouping for Global Positioning:} Trajectories are grouped into physically reachable sets reflecting the surgeon’s workflow. Lobe and capsule tracks are segmented along the channel axis with overlap; median trajectories are grouped similarly. Grouping minimizes the need for endoscope repositioning while enabling sequential execution. Global cut indices are assigned to guide ordered execution (Fig. \ref{fig:global} shows how this grouping would be conducted on an illustrated example of planned cuts). Cuts are executed 
in groups, top to bottom, with the endoscope repositioned between groups as needed.

We used a small subset of example input point clouds to program and tune our cut planner, tested its generalizability on novel inputs of new phantoms, then locked its parameters prior to the feasibility study.

\subsection{Learning Based Retraction}
\label{sec:pushcvae}

\begin{figure}[h!]
    \centering
    \includegraphics[width= 1.0\linewidth]{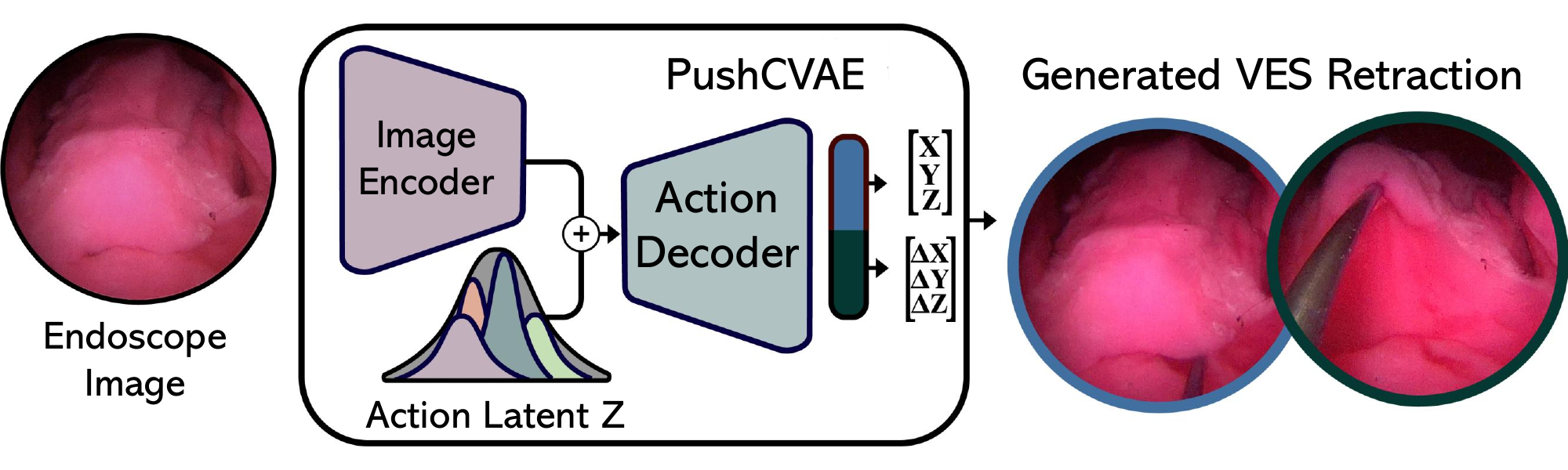}
    \caption{A simplified PushCVAE inference diagram. At runtime, the model encodes the endoscope image using a pre-trained ResNet50 backbone with a custom learned MLP to downsample the ResNet50 features to dimension 8. We then concatenate the image features to the sampled Action Latent Z and send the concatenated vector to the Action Decoder, resulting in generated push actions.}
    \label{fig:pushCVAE}
\end{figure}

For the retraction arm, we leverage a learning-based approach, PushCVAE~\cite{Watts2026_ISMR}. PushCVAE is a conditional variational autoencoder (CVAE) designed to generate soft-tissue retraction trajectories directly from monocular endoscopic images, trained from surgeon demonstrations. 
Given a single-view input image $I$, the encoder network $E_\phi$ compresses the visual scene into a latent action distribution $q_\phi(z|I)$, representing the multimodal variability of feasible retraction strategies under uncertain tissue configurations. 
The decoder $D_\theta$ samples $z \sim q_\phi(z|I)$ and predicts a corresponding three-dimensional contact point and pushing trajectory $(\hat{p}_{\text{start}}, \hat{p}_{\text{end}})$ that define a smooth, non-prehensile push motion executed by the robot’s retraction arm as outlined in Fig. \ref{fig:pushCVAE}.

We trained PushCVAE using paired and labeled demonstration data collected from expert surgeon retractions across five prostate procedures in distinct but similar phantoms. 
Each demonstration consisted of a monocular endoscopic frame and the corresponding ground-truth push trajectory obtained through the VES’s kinematic interface during surgeon teleoperation. 
Training minimizes a combined reconstruction and Kullback–Leibler (KL) divergence loss, encouraging both trajectory accuracy and a well-structured latent action space that generalizes across similar visual contexts. 
A total of 211 retraction demonstrations were recorded, evenly distributed across the three retraction phases: left trough, right trough, and middle lift. 
Due to the limited dataset size, a separate model was trained for each phase using an identical network architecture and hyperparameters to preserve consistent representation learning.

At inference, PushCVAE performs stochastic sampling to propose possible retraction actions given only the current visual frame. 
Specifically, the model draws 1000 latent samples $z_i \sim q_\phi(z|I)$ from the learned conditional Gaussian distribution and decodes each to a candidate trajectory $D_\theta(z_i, I)$. 
No external supervision, motion priors, or state feedback are provided—the model must infer the best possible retraction from visual context alone. 
Among these 1000 decoded trajectories, the action with the highest likelihood under the conditional distribution is selected and executed in real time by the robot’s retraction arm. 
In effect, the model selects its most confident prediction of a safe and effective push, given its learned uncertainty over feasible actions.

During fully autonomous median lobe removal, PushCVAE’s retraction policy operates in coordination with our model-based resection planner (see Fig.~\ref{fig:planner}). 
Phase-specific model weights (left trough, right trough, and middle lift) enable the system to maintain appropriate tissue tension and consistent visualization throughout the enucleation process without requiring any surgeon intervention.

%% file: 03-Experiments.tex
\section{Experiments and Results}

\subsection{Feasibility Study}

During an initial prototyping phase, we iteratively refined our registration workflow, testing several registration strategies through a series of trial resections and retractions before finalizing our keypoint-matching approach with fine-tuning at the start of each trough. After locking the protocol, we conducted a single-subject ($n=1$) feasibility study to evaluate system performance. Future work will involve additional tests using the same methodology to further validate and generalize the proposed framework.

\begin{figure*}[ht]
    \centering
    \includegraphics[width = \textwidth]{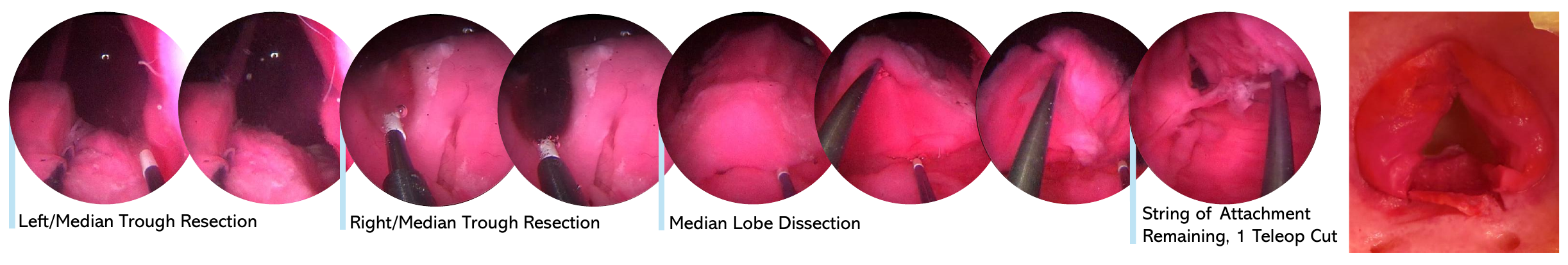}
    \caption{Intra-operative screen captures from the monocular endoscope during the feasibility study, and the final resected phantom model viewed from the proximal side for ease of visibility.}
    \label{fig:feas-process}
\end{figure*}

To begin the study, the hydrogel phantom was CT scanned and mounted near the robot. The robot was positioned within the phantom, enforcing an RCM constraint at the urethral entry. Inside the phantom, a keypoint-based registration was performed to align the segmented CT model of the prostate with the robot coordinate frame. Ten discrete keypoints were manually identified on the CT surface, selected for their distinctive feature geometry and global distribution. The keypoints were subsequently contacted in sequence using the right CTR of the VES. A keypoint-matching rigid alignment algorithm was applied to determine the optimal transformation between the CT and touched keypoints. The resulting registration matrix established a fixed spatial relationship between the CT-derived point cloud and the Kuka base frame, thereby enabling global motion of the VES relative to the phantom during subsequent phases of the procedure.

At the beginning of each of the first two surgical phases (the trough resections), global positioning of the VES was achieved via Kuka actuation, followed by a fine registration adjustment. Specifically, the CTR was used to touch the anticipated location of the first planned resection, and the positional discrepancy between the measured and planned locations was computed. A small translational correction ($<$ 3\,mm) was applied to the registration transform to compensate for error in the original registration (due to imperfect user keypoint identification and selection), as well as accumulated errors in the transformation tree during the global motion. These errors can be attributed to imperfect CAD-derived tool-to-endoscope, CTR hand-eye, and CTR-base-to-CTR-base transformations, as well as minor tissue deformations during the global repositioning.

Each surgical phase was then executed under Level 3 autonomy, with a human operator providing supervisory control. The operator issued high-level commands (either \textit{resect} or \textit{retract}). Upon receiving a \textit{resect} command, the autonomous cut planner queried the current cut number, retrieved the corresponding trajectory in the Kuka base frame, transformed it into the CTR tool frame, and executed the path at a velocity of 5\,mm/s. When a \textit{retract} command was issued, the retraction network processed the live endoscopic image to generate a distribution of appropriate push actions. The algorithm selected the maximally likely action from the distribution, unless the operator chose to select the second most likely action instead.

This process was repeated across all three surgical phases (see screen captures provided in Fig. \ref{fig:feas-process}). Phase 3 (median lobe blunt dissection) was concluded when the median lobe was fully separated from the tissue bulk. The automated procedure duration was 35 minutes. A thin residual tissue string was identified near the inferior margin of the right trough and was removed manually with one cut under teleoperation to complete the procedure. Following completion of all phases, the phantom was rescanned using CT to quantify residual median lobe volume and assess resection completeness.

\subsection{Postoperative CT Results}

Using the pre- and post-operative CT scans of the hydrogel prostate phantom, we quantify the remaining median lobe volume as a measure of procedural success, with lower volume post resection being desirable. In both CT scans, identical planar boundaries were created on the segmented lobe tissue to isolate the median lobe. The preoperative volume of the median lobe was 11.42\,cm$^3$, while the postoperative volume of residual median lobe was 2.49\,cm$^3$, indicating a percent removal of 78.2\%. To contextualize this result, we calculated what the residual median lobe volume would have been, had the planned margin been executed precisely. This target volume of residual median lobe was 2.22 cm$^3$, indicating a targeted percent removal of 80.5\%. Thus, by comparing the target percent removal to the actual result, we see that we have achieved a 97.1\% resection of the targeted median lobe volume.

\begin{figure*}[h]
    \centering
    \includegraphics[width = \textwidth]{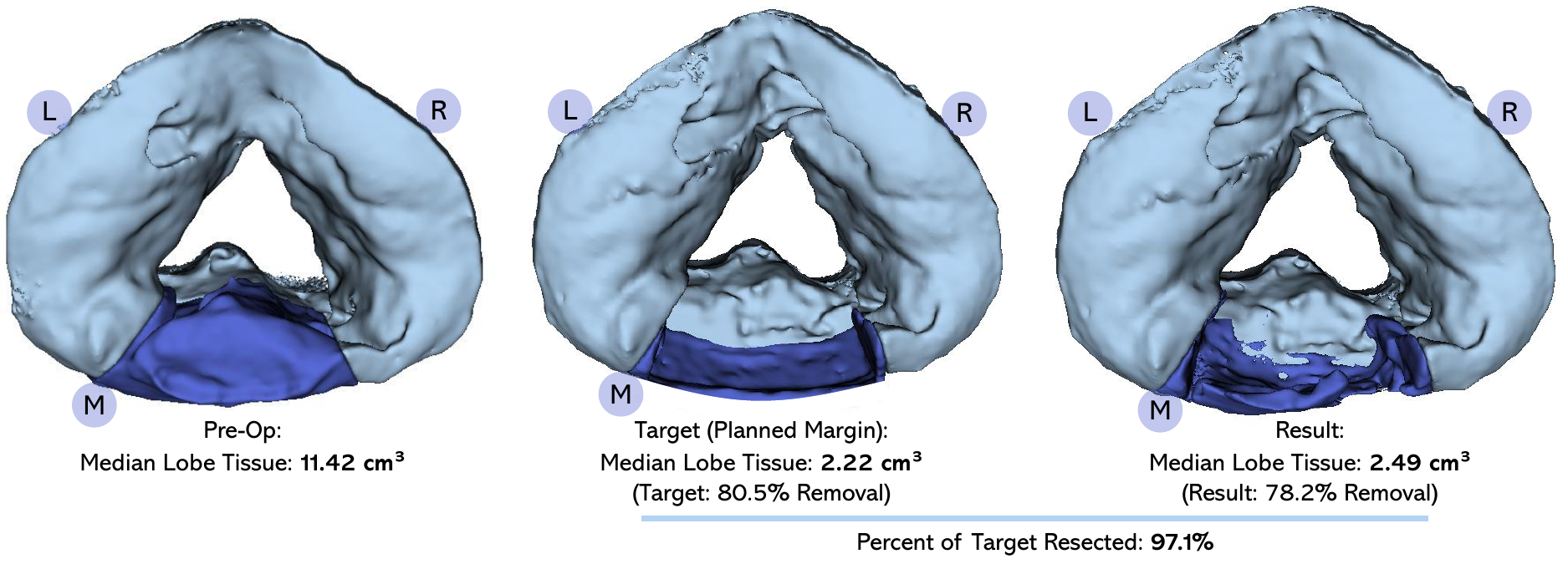}
    \caption{Pre-operative CT scan of the phantom before the feasibility study; target amount of median lobe tissue which would remain according to the planned margin; and post-operative CT scan of the phantom after the feasibility study, with corresponding results.}
    \label{fig:ctresult}
\end{figure*}

%% file: 04-Conclusion.tex
\section{Discussion and Conclusion}

This study has illustrated a first-of-its-kind procedure, showcasing a transurethral resection of the prostatic median lobe under supervised autonomy. We harness the predictability and precision of model-based control for resection, paired with the adaptability of learning-based retraction, to endow the Virtuoso Endoscopic System with surgeon-like precision in a three-phase procedural workflow. This contribution is especially significant, as it demonstrates the novel application of our fused model-based resection and learning-based retraction approach to enclosed anatomy within a minimally invasive robotic system, suggesting that automated strategies previously validated in open surgical environments can translate intuitively to minimally invasive settings.

While these findings are promising, it is important to acknowledge that this work represents an initial single-subject ($n=1$) feasibility study. Consequently, further experiments are essential to validate the reproducibility and robustness of the proposed framework. Future studies should employ the finalized registration and control protocol used in this study to ensure methodological consistency and enable a more rigorous evaluation of system performance across multiple trials.

We identify numerous drawbacks to the current method which, if improved, could yield more accurate results and greater reliability. Firstly, higher precision in calibration and registration is necessary to eliminate the fine-tuning step between phases, which currently inhibits the system's ability to achieve higher levels of autonomy. 
Additionally, the procedure is fundamentally unreliable without a form of intraoperative perception guidance for cutting.
Since prostatic lobe tissue is highly deformable, and inherently deforms during retraction, it is unreliable to plan cuts from initial static perception (CT) alone. 
Some form of updated perception, such as Simultaneous Localization and Mapping (SLAM) or Monocular Depth Estimation (MDE), must be integrated in future versions of this workflow to appropriately sense tissue deformations throughout the procedure and adjust the planned cuts accordingly. 
Closed-loop control during cutting could also be implemented via tool tracking and visual servoing, which would use the real-time endoscopic images to compensate for any snagging during cuts.

PushCVAE successfully completed the median lobe removal under full autonomy with zero surgeon intervention.
The system maintained stable tissue exposure throughout the procedure, performing 20 total autonomous retractions distributed across three distinct modes: right trough, left trough, and middle lift. 
However, PushCVAE still lacks crucial features necessary for full autonomy.
Primarily, PushCVAE is an open-loop approach that samples the highest-likelihood action from a sampled latent distribution, but does not yet incorporate uncertainty, tactile feedback, or multi-step temporal reasoning. 
Future work will focus on conditioning the latent policy on recent visual history, incorporating force cues, and enabling iterative refinement of retraction actions within each surgical phase. 
Expanding the dataset beyond the current limited number of demonstrations will enable one set of weights to learn all three modes of retraction.
Together, these advances could enable continuous, perception-driven, closed-loop autonomy for soft-tissue retraction and resection.  

Moreover, aside from the aforementioned improvements to the resection and retraction components, the transition to full autonomy is also currently inhibited by the lack of an automated state machine. Rather than the human operator prompting the system to \textit{resect} or \textit{retract}, a formal state machine could be designed to detect (i) which subtask to prompt, and (ii) subtask success via postcondition classifiers, fully replacing the human's high-level inputs. 

In summary, this study provides a proof of feasibility for autonomous transurethral lobe resection and retraction, establishing a foundation for future automation in prostate enucleation. With continued development, systems such as the VES could harness automation to broaden access to enucleation for patients with benign prostatic hyperplasia.